# *A deep learning model integrating FCNNs and CRFs for brain tumor segmentation*

*Xiaomei Zhao[1,2], Yihong Wu[1], Guidong Song[3], Zhenye Li[4], Yazhuo Zhang[3,4,5,6], and Yong Fan[7]*

[1]National Laboratory of Pattern Recognition, Institute of Automation, Chinese Academy of Sciences
[2]University of Chinese Academy of Sciences, Beijing, China
[3]Beijing Neurosurgical Institute, Capital Medical University, Beijing, China
[4]Department of Neurosurgery, Beijing Tiantan Hospital, Capital Medical University, Beijing, China
[5]Beijing Institute for Brain Disorders Brain Tumor Center, Beijing, China
[6]China National Clinical Research Center for Neurological Diseases, Beijing, China
[7]Department of Radiology, Perelman School of Medicine, University of Pennsylvania, Philadelphia, PA, USA

*Correspondence to:

Dr. Yihong Wu
National Laboratory of Pattern Recognition
Institute of Automation
Chinese Academy of Sciences
Beijing, 100190, China
Email: yhwu@nlpr.ia.ac.cn

Or

Dr. Yong Fan
Department of Radiology
Perelman School of Medicine
University of Pennsylvania
Philadelphia, PA 19104, USA
Email: yong.fan@uphs.upenn.edu; yong.fan@ieee.org
Tel: +1 215-746-4065

# ABSTRACT

Accurate and reliable brain tumor segmentation is a critical component in cancer diagnosis, treatment planning, and treatment outcome evaluation. Build upon successful deep learning techniques, a novel brain tumor segmentation method is developed by integrating fully convolutional neural networks (FCNNs) and Conditional Random Fields (CRFs) in a unified framework to obtain segmentation results with appearance and spatial consistency. We train a deep learning based segmentation model using 2D image patches and image slices in following steps: 1) training FCNNs using image patches; 2) training CRFs as Recurrent Neural Networks (CRF-RNN) using image slices with parameters of FCNNs fixed; and 3) fine-tuning the FCNNs and the CRF-RNN using image slices. Particularly, we train 3 segmentation models using 2D image patches and slices obtained in axial, coronal and sagittal views respectively, and combine them to segment brain tumors using a voting based fusion strategy. Our method could segment brain images slice-by-slice, much faster than those based on image patches. We have evaluated our method based on imaging data provided by the Multimodal Brain Tumor Image Segmentation Challenge (BRATS) 2013, BRATS 2015 and BRATS 2016. The experimental results have demonstrated that our method could build a segmentation model with Flair, T1c, and T2 scans and achieve competitive performance as those built with Flair, T1, T1c, and T2 scans.

## 1. Introduction

Accurate brain tumor segmentation is of great importance in cancer diagnosis, treatment planning, and treatment outcome evaluation. Since manual segmentation of brain tumors is laborious [1], an enormous effort has devoted to the development of semi-automatic or automatic brain tumor segmentation methods. Most of the existing brain tumor segmentation studies are focusing on gliomas that are the most common brain tumors in adults and can be measured by Magnetic Resonance Imaging (MRI) with multiple sequences, such as T2-weighted fluid attenuated inversion recovery (Flair), T1-weighted (T1), T1-weighted contrast-enhanced (T1c), and T2-weighted (T2). The segmentation of gliomas based on MRI data is challenging for following reasons: 1) gliomas may have the same appearance as gliosis and stroke in MRI data [2]; 2) gliomas may appear in any position of the brain with varied shape, appearance and size; 3) gliomas invade the surrounding brain tissues rather than displacing them, causing fuzzy boundaries [2]; and 4) intensity inhomogeneity of MRI data further increases the difficulty.

    The existing automatic and semi-automatic brain tumor segmentation methods can be broadly categorized as either generative model based or discriminative model based methods [3]. The generative model based brain tumor segmentation methods typically require prior information, which could be gained through probabilistic image atlases [4-6]. Based on probabilistic image atlases, the brain tumor segmentation problem can be modeled

as an outlier detection problem [7].

On the other hand, the discriminative model based methods solve the tumor segmentation problem in a pattern classification setting, i.e., classifying image voxels as tumor or normal tissues based on image features. The performance of discriminative model based segmentation methods are hinged on the image features and classification algorithms. A variety of image features have been adopted in tumor segmentation studies, including local histograms [8], image textures [9], structure tensor eigenvalues [10], and so on. The most commonly adopted pattern classification algorithms in brain tumor segmentation studies are support vector machines (SVMs) [11-13] and random forests [8-10, 14].

More recently, deep learning techniques have been adopted in brain tumor segmentation studies following their success in general image analysis fields, such as images classification [15], objects detection[16], and semantic segmentation [17-19]. Particularly, Convolutional Neural Networks (CNNs) were adopted for brain tumor image segmentation in the Multimodal Brain Tumor Image Segmentation Challenge (BRATS) 2014 [20-22]. More deep learning based brain tumor segmentation methods were presented in the BRATS 2015 and different deep learning models were adopted, including CNNs [23-25], convolutional restricted Boltzman machines [26], and Stacked Denoising Autoencoders [27].

Among the deep learning based tumor segmentation methods, the methods built upon CNNs have achieved better performance. Particularly, both 3D-CNNs [22, 28, 29] and 2D-CNNs [20, 21, 23-25, 30, 31] models were adopted to build tumor segmentation methods. Although 3D-CNNs can potentially take full advantage of 3D information of the MRI data, the network size and computational cost are increased too. Therefore, 2D-CNNs have been widely adopted in the brain tumor segmentation methods. Davy et al. proposed a deep learning method with two pathways of CNNs, including a convolutional pathway and a fully-connected pathway [21]. Dvorak et al. modeled the multi-class brain tumor segmentation task as 3 binary segmentation sub-tasks and each sub-task was solved using CNNs [23]. Very deep CNNs [32] were adopted to segment tumors by Pereira et al. [25]. Most of these brain tumor segmentation methods train CNNs using image patches, i.e., local regions in MR images. These methods classify each image patch into different classes, such as healthy tissue, necrosis, edema, non-enhancing core, and enhancing core. The classification result of each image patch is used to label its center voxel for achieving the tumor segmentation. Most of the above CNN brain tumor segmentation methods assumed that each voxel's label is independent, and they didn't take the appearance and spatial consistency into consideration. To take the local dependencies of labels into account, Havaei et al. constructed a cascaded architecture by taking the pixel-wise probability segmentation results obtained by CNNs trained at early stages as additional input to their following CNNs [24, 30]. To take into consideration appearance and spatial consistency of the segmentation results, Markov Random Fields (MRFs), particularly Conditional Random Fields (CRFs), have been integrated with deep learning techniques in image segmentation studies, either used as a post-process step of CNNs [28, 33] or formulated as neural networks [18, 19]. In the latter setting, both CNNs and MRFs/CRFs can be trained with back-propagation algorithms, tending to achieve better segmentation performance.

Multiple 2D CNNs could be integrated for segmenting 3D medical images. In particular, Prasoon et al. proposed a triplanar CNN [34] for knee cartilage segmentation. The triplanar network used 3 CNNs to deal with patches extracted from *xy*, *yz* and *zx* planes and fused them using a softmax classifier layer. Fritscher et al. proposed a pseudo 3D patch-based approach [35], consisting of 3 convolutional pathways for image patches in axial, coronal, and sagittal views respectively and fully connected layers for merging them. Setio et al. used multi-view convolutional networks for pulmonary nodule detection [36]. Their proposed network architecture composed multiple streams of 2D CNNs, each of which was used to deal with patches extracted in a specific angle of the nodule candidates. The outputs of the multiple streams of 2D CNNs were finally combined to detect pulmonary nodules. However, all these methods built CNNs upon image patches, not readily extendable for building FCNNs.

Preprocessing of MRI data plays an important role in the discriminative model based tumor segmentation methods that assume different MRI scans of the same modality have comparable image intensity information. The intensities of different MRI scans can be normalized by subtracting their specific mean values and dividing by their specific standard deviation values or by matching histograms [10, 22]. However, the mean values of intensities of different MRI scans do not necessarily correspond to the same brain tissue, and the histogram matching might not work well for tumor segmentation studies [8]. A robust intensity normalization has been adopted in tumor segmentation studies by subtracting the gray-value of the highest histogram bin and normalizing the standard deviation to be 1 [8].

Inspired by the success of deep learning techniques in medical image segmentation, we propose a new brain tumor segmentation method by integrating Fully Convolutional Neural Networks (FCNNs) and CRFs in a unified framework. Particularly, we formulate the CRFs as Recurrent Neural Networks [18], referred to as CRF-RNN. The integrative model of FCNNs and CRF-RNN is trained in 3 steps: 1) training FCNNs using image patches; 2) training CRF-RNN using image slices with parameters of FCNNs fixed; and 3) fine-tuning the whole network using image slices. To make use of 3D information provided by 3D medical images, we train 3 segmentation models using 2D image patches and slices obtained in axial, coronal and sagittal views respectively, and combine them to segment brain tumors using a voting based fusion strategy. The proposed method is able to segment brain images slice-by-slice, which is much faster than the image patch based segmentation methods. Our method could achieve competitive segmentation performance based on 3 MR imaging modalities (Flair, T1c, T2), rather than 4 modalities (Flair, T1, T1c, T2) [3, 8-10, 14, 20-31], which could help reduce the cost of data acquisition and storage. We have evaluated our method based on imaging data provided by the Multimodal Brain Tumor Image Segmentation Challenge (BRATS) 2013, the BRATS 2015, and the BRATS 2016. The experimental results have demonstrated that our method could achieve promising brain tumor segmentation performance. Preliminary results have been reported in a conference proceeding paper of the BRATS 2016 [37].

## 2. Methods and materials

### 2.1 Imaging data

All the imaging data used in this study were obtained from the BRATS 2013[2], the BRATS 2015[3] and the BRATS 2016[4]. The BRATS 2013 provided clinical imaging data of 65 glioma patients, including 14 patients with low-grade gliomas (LGG) and 51 patients with high-grade gliomas (HGG). The patients were scanned with MRI scanners from different vendors at 4 different centers, including Bern University, Debrecen University, Heidelberg University, and Massachusetts General Hospital. Each patient had multi-parametric MRI scans, including T2-weighted fluid attenuated inversion recovery (Flair), T1-weighted (T1), T1-weighted contrast-enhanced (T1c), and T2-weighted (T2). All the MRI scans of the same patient were rigidly co-registered to their T1c scan and resampled at 1 mm isotropic resolution in a standardized axial orientation with a linear interpolator [3]. All images were skull stripped. Ground truths were produced by manual annotations. The cases were split into training and testing sets. The training set consists of 20 HGG and 10 LGG cases. The testing set consists of Challenge and Leaderboard subsets. The Challenge dataset has 10 HGG cases and the Leaderboard dataset contains 21 HGG and 4 LGG.

The imaging dataset provided by BRATS 2015 contains imaging data obtained from the BRATS 2012, 2013, and the NIH Cancer Imaging Archive (TCIA). Each case has Flair, T1, T1c, and T2 scans aligned onto the same anatomical template space and interpolated at 1 mm$^3$ voxel resolution. The testing dataset consists of 110 cases with unknown grades, and the training dataset consists of 220 HGG and 54 LGG cases. In the testing dataset, the ground truth of each case was produced by manual annotation. In the training dataset, all the cases from the BRATS 2012 and 2013 were labeled manually, and the cases from the TCIA were annotated by fusing segmentation results obtained using top-ranked methods of the BRATS 2012 and 2013. The annotations were inspected visually and approved by experienced raters. The tumor labels of the training cases are provided along with their imaging scans, while only imaging data are provided for the testing cases for blind evaluation of the segmentation results.

BRATS 2016 shares the same training dataset with BRATS 2015, which consists of 220 HGG and 54 LGG. Its testing dataset consists of 191 cases with unknown grades. The ground truth of each testing case was produced by manual annotation, not released to the competition participants.

### 2.2 Brain tumor segmentation methods based on FCNNs trained using image patches

Deep learning techniques, particularly CNNs, have been successfully adopted in image segmentation studies. A deep learning model of CNNs usually has millions or even billions of parameters. To train the deep CNNs with sufficient training samples, image patch-based techniques are adopted [20-25, 28, 30, 31, 38-40]. With the image patch based representation, the image segmentation problem can be solved as a classification problem of image patches.

---

[2] https://www.virtualskeleton.ch/BRATS/Start2013
[3] https://www.virtualskeleton.ch/BRATS/Start2015
[4] https://www.virtualskeleton.ch/BRATS/Start2016

An image patch is a local region extracted from an image to characterize its central pixel/voxel in 2D/3D, and has the same label as its center pixel/voxel's label in the classification problem. In the training phase, a large number of image patches can be extracted to train the CNNs. In the testing phase, image patches extracted from a testing image are classified one by one by the trained CNNs. Then, the classification results of all image patches make up a segmentation result of the testing image. However, FCNNs can segment a testing image slice by slice with improved computational efficiency [30], even though the model is trained using image patches. Since the number and location of training image patches for each class can be easily controlled by changing the image patch sampling scheme, image patch-based deep learning segmentation methods can avoid the training sample imbalance problem. However, a limitation of image patch-based segmentation methods is that relationship among image patches is typically lost. Integrating CRF-RNN with FCNNs tends to overcome such a limitation in tumor segmentation.

## 2.3 The proposed brain tumor segmentation method

The proposed brain tumor segmentation method consists of 4 main steps: pre-processing, segmenting image slices using deep learning models with integrated FCNNs and CRF-RNN from axial, coronal and sagittal views respectively, fusing segmentation results obtained in the three different views, and post-processing.

### 2.3.1 Pre-processing of the imaging data

Since MRI scans typically have varied intensity ranges and are affected by bias fields differently, we adopted a robust intensity normalization method to make MRI scans of different patients comparable, besides correcting the bias field of MRI data using N4ITK [41]. Our normalization method is built upon the image mode based method [8], which normalizes image intensity by subtracting the image mode (e.g. the gray-value of the highest histogram bin) and normalizing the standard deviation to be 1. As almost half of the brain is the whiter matter [42], the gray-value of the highest histogram bin typically corresponds to the gray-value of the white matter, and therefore matching intensity values of the white matter across MRI scans and normalizing the intensity distributions accordingly would largely make different MRI scans comparable. However, the standard deviation calculated based on intensity mean value does not necessarily have a fixed tissue meaning. Therefore, in our study a robust intensity deviation is adopted to replace the standard deviation used in [8]. The robust deviation is computed based on the gray-value of the highest histogram bin, representing the discreteness of intensity to the gray-value of white matter. Besides, the intensity mean is more sensitive to noise than the gray value of the highest histogram bin. Thus the standard deviation calculated based on intensity mean is more sensitive to noise than the robust deviation.

Given an MRI scan $V$ with voxels $\{v_1, v_2, \cdots, v_N\}$, and each voxel $v_k$ has intensity $I_k$, $k = 1, 2, \cdots, N$, the robust deviation $\tilde{\sigma} = \sqrt{\sum_{k=1}^{N}(\hat{I}-I_k)^2/N}$, where $\hat{I}$ denotes the gray-value of the highest histogram bin. Our intensity normalization procedure is following:

Step 1. Transform the intensity range to 0-255 linearly.

Step 2. Calculate the intensity histogram, with 256 bins.

Step 3. Subtract the gray-value of the highest histogram bin $\hat{I}$ and divide the robust deviation .

Step 4. Multiply each voxel's intensity by a constant σ and plus a constant $I_0$. Then, set the intensities that are below 0 or above 255 to 0 and 255 respectively. In the present study, we set σ and $I_0$ equal to the gray-value of the highest histogram bin and robust deviation of NO. 0001 HGG clinical training image data of BRATS 2013, which has been pre-processed by N4ITK and the step 1. For the Flair, T1c, and T2 scans, σ=30, 31, 37 and $I_0$=75, 99, 55 respectively.

The image intensity normalization effect is illustrated with T2 scans in Fig. 1. Particularly, we randomly selected 3 subjects from the BRATS 2013 and 3 subjects from the BRATS 2015. The results shown in Fig. 1 clearly demonstrated that the image intention normalization could improve comparability of different scans. The improvement is further confirmed by image intensity histograms of 30 subjects from the BRATS 2013 training dataset, as shown in Fig.2.

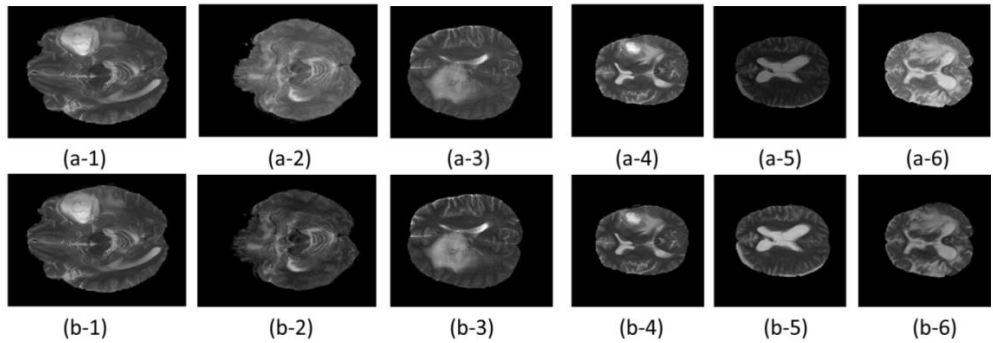

Fig.1. T2 scans before (top row) and after (bottom row) the proposed intensity normalization. (a-1)-(a-3) and (b-1)-(b-3) are randomly selected subjects from the BRATS 2013, and (a-4)-(a-6) and (b-4)-(b-6) are randomly selected subjects from the BRATS 2015. (a-1)-(a-6): before normalization; (b-1)-(b-6): after normalization. All the scans were preprocessed by N4ITK and the proposed normalization step 1.

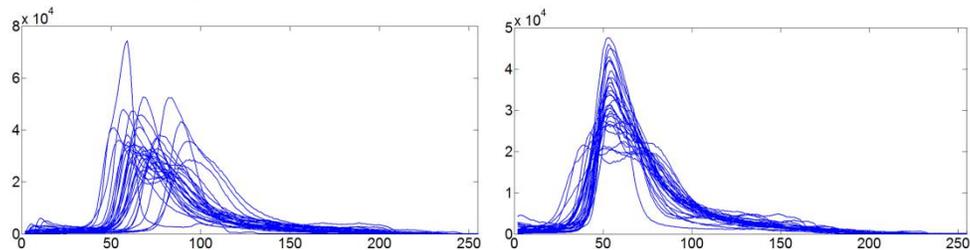

Fig.2. Image intensity histograms of T2 scans of 30 subjects from the BRATS 2013 training dataset before (left) and after (right) the intensity normalization. All the scans were preprocessed by N4ITK and the proposed normalization step 1.

### 2.3.2 A deep learning model integrating FCNNs and CRFs

The proposed deep learning model for brain tumor segmentation integrates Fully Convolutional Neural Networks

(FCNNs) and Conditional Random Fields (CRFs), as illustrated by Fig. 3. We formulated CRFs as Recurrent Neural Networks (RNNs), referred to as CRF-RNN [18]. The proposed method could segment brain images slice by slice.

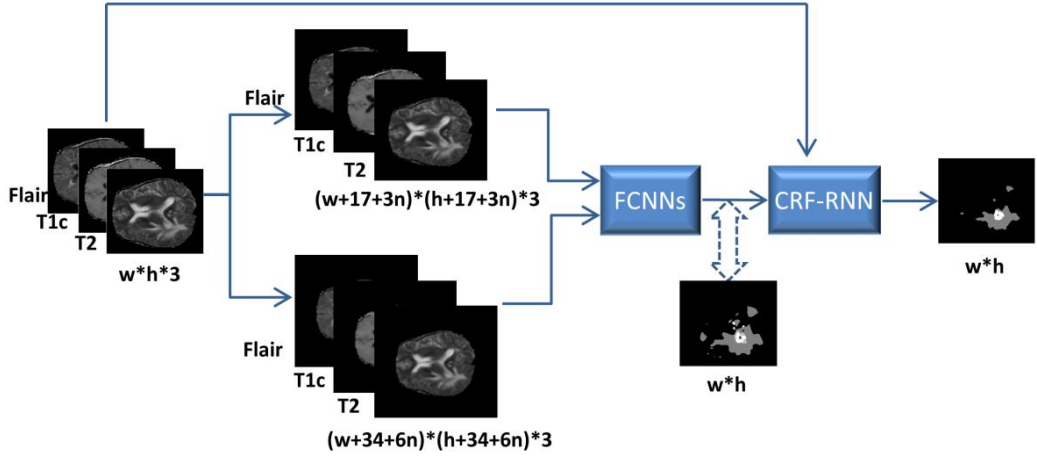

Fig.3. Flowchart of the proposed deep learning model integrating FCNNs and CRFs for brain tumor segmentation.

## (1) FCNNs

The structure of our proposed FCNNs is illustrated by Fig. 4. Similar to the network architectures proposed in [28, 30], the inputs to our network are also in 2 different sizes. Passing through a series of convolutional and pooling layers, the larger inputs turn into feature maps with the same size of smaller inputs. These feature maps and smaller inputs are sent into following networks together. In this way, both local image information and context information in a larger scale can be taken into consideration for classifying image patches. Different from the cascaded architecture proposed in [30], the two branches in our FCNNs are trained simultaneously, rather than trained in different steps. Furthermore, our model has more convolutional layers.

Our deep FCNNs are trained using image patches, which are extracted from slices of the axial view, coronal view or sagittal views randomly. Equal numbers of training samples for different classes are extracted to avoid data imbalance problem. There are 5 classes in total, including healthy tissue, necrosis, edema, non-enhancing core, and enhancing core.

As shown in Fig. 4, in our deep FCNNs, the kernel size of each max pooling layer is set to $n \times n$, and the size of image patches used to train FCNNs is proportional to the kernel size. Different settings of the kernel size or equivalently the image patch size may affect the tumor segmentation performance. The max pooling layers of our FCNNs are used to capture image information in large scales with a relatively small number of network parameters. We set the stride of each layer to be 1. Therefore, in the testing stage, our model can segment brain images slice by slice.

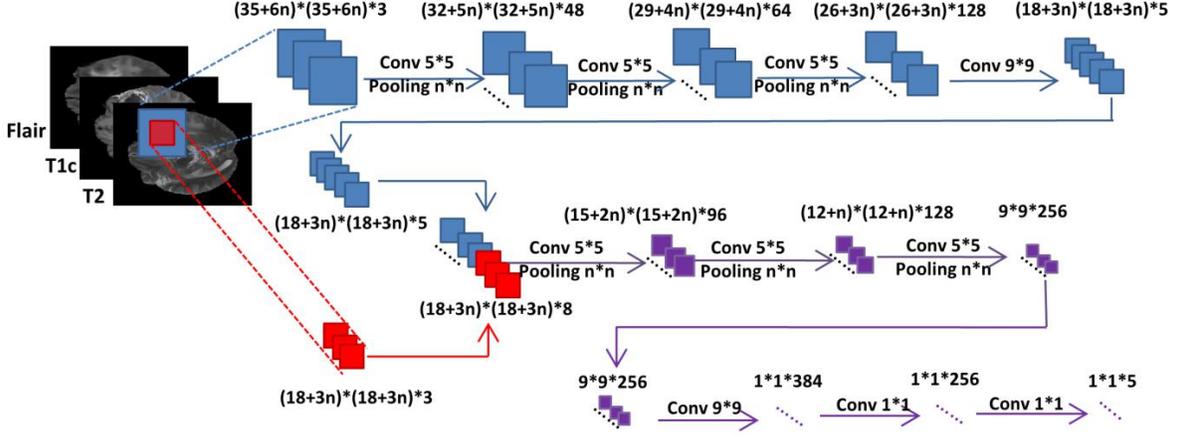
Fig.4. The network structure of our deep FCNNs.

*(2) CRF-RNN*

CRF-RNN formulates 2D fully connected Conditional Random Fields as Recurrent Neural Networks [18]. Given a 2D image $I$, comprising a set of pixels $\{I_i | i = 1, \cdots, M\}$, the image segmentation problem is solved as an optimization problem using fully connected CRFs by minimizing an energy function [43]:

$$E(Y) = \sum_{i=1}^{M} \Phi(y_i^u) + \sum_{\forall i,j, i<j} \Psi(y_i^u, y_j^v), \tag{1}$$

where $Y$ is a certain label assignment to $I$, $i,j \in \{1, \cdots, M\}$, $y_i^u$ denotes the assignment of label $u$ to pixel $I_i$, $y_j^v$ denotes the assignment of label $v$ to pixel $I_j$, $u,v \in L = \{l_1, l_2, \cdots, l_C\}$ are segmentation labels, the unary term $\Phi(y_i^u)$ measures the cost of assigning label $u$ to pixel $I_i$, and the pairwise term $\Psi(y_i^u, y_j^v)$ measures the cost of assigning label $u$ and $v$ jointly to $I_i$ and $I_j$. According to [19], minimizing $E(Y)$ equals to minimizing an energy function:

$$F(Q) = \sum_{\forall i} \sum_{\forall u \in L} q_i^u \Phi(y_i^u) + \sum_{\forall i,j, i<j} \sum_{\forall u \in L} \sum_{\forall v \in L} q_i^u q_j^v \Psi(y_i^u, y_j^v) + \sum_{\forall i} \sum_{\forall u \in L} q_i^u \ln q_i^u, \tag{2}$$

where $q_i^u$ denotes the probability of assigning label $u$ to pixel $I_i$, which is the variable that we aim to estimate.

Differentiating Eqn. (2) with respect to $q_i^u$ and setting the differentiation result equal to 0, we have

$$q_i^u \propto \exp\{-\Phi(y_i^u) - \sum_{j, j \neq i} \sum_{\forall v \in L} q_j^v \Psi(y_i^u, y_j^v)\}, \tag{3}$$

The unary term $\Phi(y_i^u)$ can be obtained from the FCNNs, and the pairwise potential $\Psi(y_i^u, y_j^v)$ is defined as

$$\Psi(y_i^u, y_j^v) = \mu(u,v) \sum_{m=1}^{K} w^{(m)} k^{(m)}(f_i, f_j), \tag{4}$$

where $K = 2$ is the number of Gaussian kernel; $k^{(m)}$ is a Gaussian kernel, $k^{(1)} = \exp\left(-\frac{|s_i - s_j|}{2\theta_\alpha^2} - \frac{|e_i - e_j|}{2\theta_\beta^2}\right)$ and $k^{(2)} = \exp\left(-\frac{|s_i - s_j|}{2\theta_\gamma^2}\right)$ ($e_i$ and $e_j$ denote the intensity of $I_i$ and $I_j$ respectively, $s_i$ and $s_j$ denote spatial coordinates of $I_i$ and $I_j$, $\theta_\alpha$, $\theta_\beta$ and $\theta_\gamma$ are parameters of the Gaussian kernels); $w^{(m)}$ is a weight for the

Gaussian kernel $k^{(m)}$; $f_i$ and $f_j$ denote image feature vectors of $I_i$ and $I_j$ respectively, encoding their intensity ($e_i$, $e_j$) and spatial position information ($s_i$, $s_j$); $\mu(u,v)$ indicates the compatibility of labels $u$ and $v$. Substituting (4) into (3), we get:

$$q_i^u \propto exp\{-\Phi(y_i^u) - \sum_{\forall v \in L}\mu(u,v)\sum_{m=1}^{K} w^{(m)} \sum_{\forall j \neq i} k^{(m)}(f_i,f_j)q_j^v\}. \tag{5}$$

Fully connected CRF predicts the probability of assigning label $u$ to pixel $I_i$ according to Eqn. (5), and $q_i^u$ can be calculated using a mean field iteration algorithm formulated as Recurrent Neural Networks so that CNNs and the fully connected CRF are integrated as one deep network and can be trained using a back-propagation algorithm [18]. Fig. 5 shows the network structure of CRF-RNN. G1 and G2 in Fig. 5 are two gating functions:

$$Q_{in} = \begin{cases} P_{norm} = softmax(P), \ initialization, t = 0 \\ Q_{out} = one\ meanfield\ interation(Q_{in}), \ 0 < t \leq T \end{cases}, \tag{6}$$

$$Q_{final} = \begin{cases} 0, \ 0 < t < T \\ Q_{out}, \ t = T \end{cases}, \tag{7}$$

where $Q = \{q_i^u | \forall i \in [1,2,\cdots,M], \forall u \in L\}$, $Q_{in}$ denotes the input $Q$ of one mean-field iteration; $Q_{out}$ denotes the output $Q$ of one mean-field iteration; $Q_{final}$ denotes the final prediction results of CRF-RNN. $P$ denotes the output of FCNNs, and $P_{norm}$ denotes the $P$ that after softmax operation. $t$ represents the $t^{th}$ mean-field iteration, and $T$ is the total number of mean-field iterations. In our study, the unary term $-\Phi(y_i^u)$ is the output of FCNNs, and the pairwise potential $\Psi(y_i^u, y_j^v)$ are computed based on pixel features $f_i$ and $f_j$ with information provided by Flair, T1c and T2 slices with $\theta_\alpha = 160$, $\theta_\beta = 3$, $\theta_\gamma = 3$ while $w$ and $\mu$ are learned in the training phase [18]. By integrating FCNNs and CRF-RNN in one deep network, we are able to train the network end-to-end with a typical back-propagation algorithm [18].

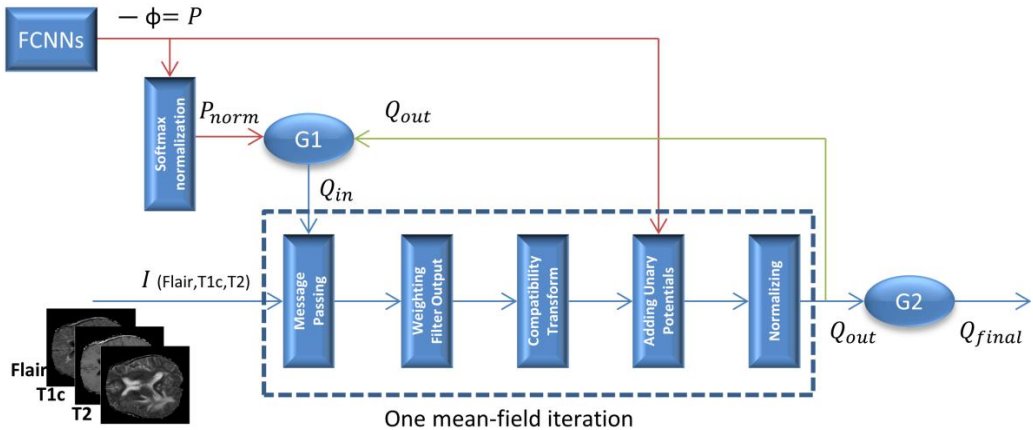

Fig.5. The network structure of CRF-RNN.

*(3) The integration of FCNNs and CRF-RNN*

The proposed brain tumor segmentation network consists of FCNNs and CRF-RNN. The FCNNs predict the probability of assigning segmentation labels to each pixel, and the CRF-RNN takes the prediction results and image information as its input to globally optimize appearance and spatial consistency of the segmentation results according to each pixel's intensity and position information.

The proposed deep learning network of FCNNs and CRF-RNN is trained in 3 steps: 1) training FCNNs using image patches; 2) training CRF-RNN using image slices with parameters of FCNNs fixed; and 3) fine-tuning the whole network using image slices.

Once the fine-tune of deep learning based segmentation model is done, the model can be applied to image slices one by one for segmenting tumors. Given an $w \times h$ image slice with 3 channels, i.e., pre-processed Flair, T1c, and T2 scans respectively, we first pad the image slice with zeros to create 2 larger images with sizes of $(w + 17 + 3n) \times (h + 17 + 3n) \times 3$ and $(w + 34 + 6n) \times (h + 34 + 6n) \times 3$ respectively. Using these 2 larger images as inputs of the FCNNs, we obtain 5 label predication images $P^u, u = 1,2,3,4,5$, $P^u = \{p_{i,j}^u | i \in [1,2,...,w], j \in [1,2,...,h]\}$, with the same size of the original image slices. $p_{i,j}^u$ represents one pixel's predicted probability of brain tissue labels, such as healthy tissue, necrosis, edema, non-enhancing core or enhancing core. Then, these label predication images $P = \{P^u | u = 1,2,3,4,5\}$ along with the image slice $I = \{I_{Flair}, I_{T1c}, I_{T2}\}$ are used as inputs to the CRF-RNN. Finally, the CRF-RNN obtains a globally optimized segmentation result of the original image slice. Fig. 3 shows the flowchart of the proposed deep learning model integrating FCNNs and CRF-RNN for brain tumor segmentation.

In the training steps 2 and 3, we first calculate softmax loss according to the current segmentation results and the ground truth, and then the loss information is back-propagated to adjust network parameters of the integrated FCNNs and CRF-RNN. In the training step 2, we fix FCNNs and adjust the parameters in CRF-RNN. In the training step 3, we set a small learning rate and fine-tune the parameters of the whole network. In our experiments, the initial learning rate was set to $10^{-5}$ and the learning rate was divided by 10 after each 20 epoches in the training step 1, and the learning rate was set to $10^{-8}$ and $10^{-10}$ respectively in the training steps 2 and 3.

*2.3.3 Fusing segmentation results obtained in axial, coronal and sagittal views*

We train 3 segmentation models using patches and slices of axial, coronal and sagittal views respectively. During testing, we use these 3 models to segment brain images slice by slice in 3 different views, yielding 3 segmentation results. A majority voting strategy is adopted to fuse the segmentation results. Let $r_a$, $r_c$, and $r_s$ denote the segmentation results of one voxel gotten in axial, coronal and sagittal views respectively, let $r$ denote the segmentation result after fusion, let 0,1,2,3,4 denote a voxel labeled as healthy tissue, necrosis, edema, non-enhancing core, and enhancing core respectively, the fused segmentation result is obtained by following voting procedure:

Step 1. If two or more than two of $r_a$, $r_c$, and $r_s$ are above 0, let $r = 2$.
Step 2. If two or more than two of $r_a$, $r_c$, and $r_s$ equal to 1, let $r = 1$.

Step 3. If two or more than two of $r_a$, $r_c$, and $r_s$ equal to 3, let $r = 3$.

Step 4. If two or more than two of $r_a$, $r_c$, and $r_s$ equal to 4, let $r = 4$.

### 2.3.4 Post-processing

To further improve the brain tumor segmentation performance, a post-processing method is proposed. Hereinafter, $V_{Flair}$, $V_{T1c}$, $V_{T2}$ denote pre-processed Flair, T1c, T2 MR images respectively, $Res$ denotes the segmentation result obtained by our integrated deep learning model, $V_{Flair}(x,y,z)$, $V_{T1c}(x,y,z)$, $V_{T2}(x,y,z)$, and $Res(x,y,z)$ denote the value of voxel $(x,y,z)$ in $V_{Flair}$, $V_{T1c}$, $V_{T2}$, and $Res$ respectively, $Res(x,y,z) = 0,1,2,3,4$ indicates that the voxel $(x,y,z)$ is labeled as healthy tissue, necrosis, edema, non-enhancing core, and enhancing core respectively, $Mean_{Flair}$ and $Mean_{T2}$ denote the average intensity of the whole tumor region indicated by $Res$ in $V_{Flair}$ and $V_{T2}$ scans. For a segmentation result $Res$ with $N$ 3D connected tumor regions, $mean_{flair}(n)$ and $mean_{t2}(n)$ denote the average intensity of the $n^{th}$ 3D connected tumor area in $V_{Flair}$ and $V_{T2}$ respectively. The post-processing method consists of following step:

Step 1. If $mean_{flair}(n) > \theta_{11}$ and $mean_{t2}(n) > \theta_{12}$, set all voxels in the $n^{th}$ 3D connected tumor area in $Res$ to be 0 so that the $n^{th}$ 3D connected tumor region is removed from $Res$, taking into consideration that isolated local areas with super high intensities are usually caused by imaging noise rather than tumors. In the present study, $\theta_{11} = \theta_{12} = 150$.

Step 2. If a voxel $(x,y,z)$ satisfies the following conditions at the same time:

①$V_{Flair}(x,y,z) < \theta_{21} \times Mean_{Flair}$, ②$V_{T1c}(x,y,z) < \theta_{22}$, ③$V_{T2}(x,y,z) < \theta_{23} \times Mean_{T2}$, ④$Res(x,y,z) < 4$, set $Res(x,y,z) = 0$.

In general, tumor tissues have high signal in at least one modality of Flair, T1c, and T2. Voxels with low signal in Flair, T1c, and T2 at the same time are generally not tumor tissues. Thus, this step removes those segmented tumor regions whose intensities in Flair, T1c, T2 are below 3 thresholds respectively. However, enhancing core is an exception. In the present study, $\theta_{21} = 0.8$, $\theta_{22} = 125$, $\theta_{23} = 0.9$.

Step 3. Let $volume(n)$ denote the volume of the $n^{th}$ 3D connected tumor area in $Res$. $Volume_{max}$ is the volume of the maximum 3D connected tumor area in $Res$. If $volume(n)/Volume_{max} < \theta_{31}$, remove the $n^{th}$ 3D connected segmented tumor region in $Res$. In the present study, $\theta_{31} = 0.1$.

Step 4. Fill the holes in $Res$ with necrosis. Holes in $Res$ are very likely to be necrosis.

Step 5. If $V_{T1c}(x,y,z) < \theta_{41}$ and $Res(x,y,z) = 4$, set $Res(x,y,z) = 1$. Our model may mistakenly label necrosis areas as enhancing core. This step corrects this potential mistake through a threshold in T1c. In the present study, $\theta_{41} = 100$.

Step 6. Let $vol_e$ denote the volume of enhancing core represented in $Res$, and $vol_t$ denote the volume of the whole tumor. If $vol_e/vol_t < \theta_{61}$, $V_{T1c}(x,y,z) < \theta_{62}$, and $Res(x,y,z) = 2$, set $Res(x,y,z) = 3$. Our tumor segmentation model is not sensitive to non-enhancing core. In our model, non-enhancing regions might be mistakenly labeled as edema, especially when the enhancing core region is very small. In the present study, $\theta_{61} = 0.05$, $\theta_{62} = 85$.

The parameters were set based on the BRATS 2013 dataset. Since the number of training cases of the BRATS 2013 is small, we did not cross-validate the parameters, therefore they are not necessarily optimal. We used the same parameters in all of our experiments, including our experiments on BRATS 2013, 2015 and 2016. In addition to the aforementioned post-processing steps, we could also directly use CRF as a post-processing step of FCNNs as did in a recent study [28].

## 3. Experiments

Our experiments were carried out based on imaging data provided by the BRATS 2013, 2015 and 2016 on a computing server with multiple Tesla K80 GPUs and Intel E5-2620 CPUs. However, only one GPU and one CPU were useable at the same time for our experiments. Our deep learning models were built upon Caffe [44].

Based on the BRATS 2013 data, a series of experiments were carried out to evaluate how different implementation of the proposed method affect tumor segmentation results with respect to CRF, post-processing, image patch size, number of training image patches, pre-processing, and imaging scans used. We also present segmentation results obtained for the BRATS 2013. The segmentation model was built upon the training data and then evaluated based on the testing data. Since no ground truth segmentation result for the testing data was provided, all the segmentation results were evaluated by the BRATS evaluation website. The tumor segmentation performance was evaluated using the BRATS segmentation evaluation metrics for complete tumor, core region, and enhancing region, including Dice, Positive Predictive Value (PPV), and Sensitivity. Particularly, the complete tumor includes necrosis, edema, non-enhancing core, and enhancing core; the core region includes necrosis, non-enhancing core, and enhancing core; and the enhancing region only includes the enhancing core. The tumor segmentation evaluation metrics are defined as follows:

$$Dice(P_*,T_*) = \frac{|P_* \cap T_*|}{(|P_*|+|T_*|)/2}, PPV(P_*,T_*) = \frac{|P_* \cap T_*|}{|P_*|}, Sensitivity(P_*,T_*) = \frac{|P_* \cap T_*|}{|T_*|},$$

where $*$ indicates complete, core or enhancing region, $T_*$ denotes the manually labeled region, $P_*$ denotes the segmented region, $|P_* \cap T_*|$ denotes the overlap area between $P_*$ and $T_*$, and $|P_*|$ and $|T_*|$ denote the areas of $P_*$ and $T_*$ respectively.

### 3.1. Experiments on BRATS 2013 dataset

The BRATS 2013 training dataset contains 10 LGG and 20 HGG. Its testing dataset consists of two subsets, namely Challenge and Leaderboard. The Challenge dataset has 10 HGG cases and the Leaderboard dataset contains 21 HGG and 4 LGG.

A number of experiments were carried out based on the BRATS 2013 dataset, including 1) comparing the segmentation performance of FCNNs with and without post-processing, and the performance of the proposed deep learning network integrating FCNNs and CRF-RNN (hereinafter referred to as FCNN+CRF) with and without post-processing, in order to validate the effectiveness of CRFs and post-processing; 2) evaluating the segmentation performance of FCNN+CRF with 5 post-processing steps (6 post-processing steps in total), in order to test the effectiveness of each post-processing step; 3) evaluating the segmentation performance of FCNNs trained using different sizes of patches; 4) evaluating the segmentation performance of FCNNs trained using different numbers of patches; 5) comparing the segmentation performance of segmentation models built upon scans of 4 imaging sequences (Flair, T1, T1c, and T2) and 3 imaging sequences (Flair, T1c, and T2); and 6) evaluating how the image preprocessing step affect the segmentation performance. All the above experiments were performed in axial view. Apart from these experiments described above, we show the effectiveness of fusing segmentation results of three views in Section 3.1.7 and summarize comparison results with other methods in Section 3.1.8.

### 3.1.1. Evaluating the effectiveness of CRFs and post-processing

Table 1 shows the evaluation results of FCNNs with and without post-processing, and FCNN+CRF (our integrated network of FCNNs and CRF-RNN) with and without post-processing on the BRATS 2013 Challenge dataset and Leaderboard dataset. These results demonstrated that CRFs improved the segmentation accuracy and so did the post-processing. With respect to both Dice and PPV, FCNN+post-process and FCNN+CRF improved the segmentation performance in all complete tumor, core region, and enhancing region. However, CRFs and post-process reduced Sensitivity. It is worth noting that CRFs improved Sensitivity of the enhancing region. In summary, CRFs improved both the Dice and PPV and decreased the Sensitivity on the complete and core regions, FCNN+CRF+post-process obtained the best performance with respect to Dice and PPV, but degraded the performance with respect to the Sensitivity, especially on the complete tumor region.

We also adopted a 3D CRF based post-processing step as did in a recent study [28]. Particularly, the parameters of the 3D CRF were optimized by grid searching based on the training dataset of BRATS 2013. Table 1 summarizes segmentation scores obtained by our method with different settings. These results indicated that 3D CRF as a post-processing step could improve the segmentation performance as 3D information was taken into consideration. However, our proposed post-processing procedure could further improve the segmentation performance.

Fig. 6 shows representative segmentation results on the BRATS 2013 Challenge dataset. These segmentation results demonstrated that FCNN+CRF could improve the spatial and appearance consistence of segmentation results, and FCNN+CRF+post-process could reduce false positives.

Table 1. Evaluation results of FCNNs with and without post-processing, FCNN+CRF with and without post-processing, and FCNN+3D-CRF with and without post-processing. (The sizes of image patches used to train FCNNs were 33*33*3 and 65*65*3 respectively, *n*=5, and the number of patches used to train FCNNs was 5000*5*20. FCNN+CRF is short for the integrated network of FCNNs and CRF-RNN)

| Dataset | Methods | Dice | | | PPV | | | Sensitivity | | |
|---|---|---|---|---|---|---|---|---|---|---|
| | | complete | core | enhancing | complete | core | enhancing | complete | core | enhancing |
| Challenge | FCNNs | 0.74 | 0.72 | 0.67 | 0.62 | 0.63 | 0.60 | **0.94** | **0.86** | 0.77 |
| | FCNN+post-process | 0.81 | 0.75 | 0.73 | 0.73 | 0.70 | 0.73 | **0.94** | 0.85 | 0.74 |
| | FCNN+CRF | 0.85 | 0.80 | 0.70 | 0.87 | 0.80 | 0.63 | 0.84 | 0.81 | **0.80** |
| | FCNN+CRF+post-process | **0.87** | **0.83** | 0.76 | **0.92** | **0.87** | 0.77 | 0.83 | 0.81 | 0.77 |
| | FCNN+3D-CRF | 0.85 | 0.80 | 0.73 | 0.84 | 0.78 | 0.69 | 0.88 | 0.83 | **0.80** |
| | FCNN+3D-CRF +post-process | **0.87** | **0.83** | **0.78** | 0.89 | 0.85 | **0.78** | 0.86 | 0.83 | 0.79 |
| Leaderboard | FCNNs | 0.70 | 0.61 | 0.54 | 0.58 | 0.57 | 0.49 | **0.96** | 0.74 | 0.67 |
| | FCNN+post-process | 0.81 | 0.65 | 0.61 | 0.74 | 0.63 | 0.62 | 0.94 | 0.78 | 0.66 |
| | FCNN+CRF | 0.83 | 0.66 | 0.57 | 0.85 | 0.71 | 0.50 | 0.85 | 0.69 | **0.71** |
| | FCNN+CRF+post-process | 0.86 | **0.73** | 0.62 | **0.89** | 0.76 | **0.64** | 0.84 | 0.78 | 0.68 |
| | FCNN+3D-CRF | 0.84 | 0.65 | 0.61 | 0.81 | 0.71 | 0.57 | 0.90 | 0.71 | **0.71** |
| | FCNN+3D-CRF +post-process | **0.87** | 0.71 | **0.63** | 0.88 | 0.74 | 0.63 | 0.88 | **0.79** | 0.70 |

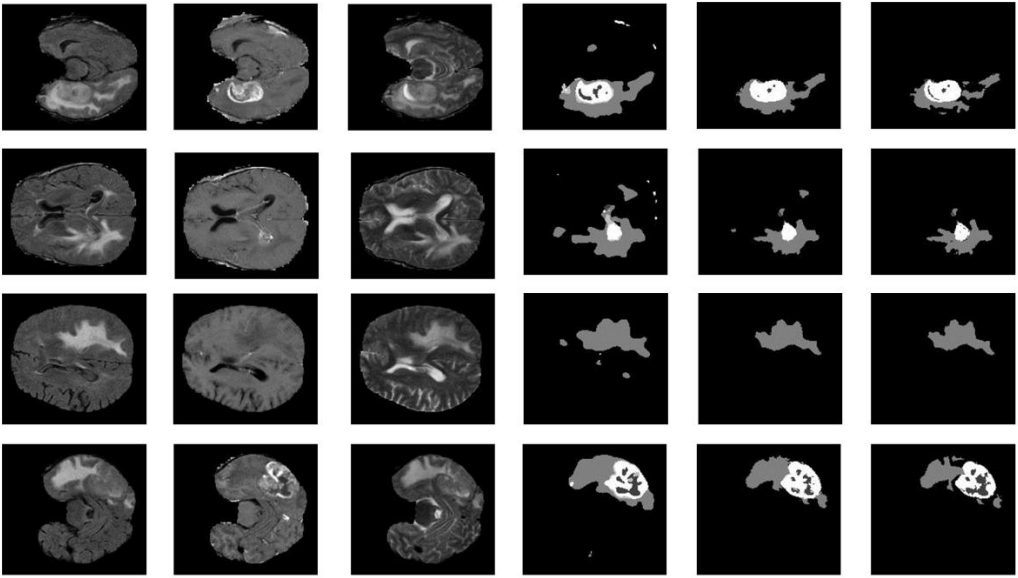

Fig.6. Example segmentation results on the BRATS 2013 Challenge dataset. The first and second rows show the segmentation results of the 50th and 80th slice of the axial view of Subject 0301. The third and fourth rows show the segmentation results of the 40th and 70th slice of the axial view of Subject 0308. From left to right: Flair, T1c, T2, segmentation results of FCNNs, segmentation results of FCNN+CRF, and segmentation results of FCNN+CRF+post-process. In the segmentation results, each gray level represents a tumor class, from low to high: necrosis, edema, non-enhancing core, and enhancing core.

### 3.1.2. Evaluating the effectiveness of each post-processing step

To investigate the effectiveness of each post-processing step, we obtained segmentation results of FCNN+CRF+post-$x$, $x$=1, 2, …, 6. In particular, FCNN+CRF+post-$x$ indicates FCNN+CRF with all other post-processing steps except the step $x$. As described in section 2.3.4, our post-processing consists of 6 steps in total. All the evaluation results are summarized in Table 2. These results indicated that the post-processing step 3 played the most important role in the tumor segmentation, although all these post-processing steps might contribute the segmentation.

Table 2. The evaluation results of FCNN+CRF with 5 of 6 post-processing steps (FCNN+CRF+post-$x$ indicates FCNN+CRF with all other post-processing steps except the step $x$, the sizes of patches used to train FCNNs were 33*33*3 and 65*65*3 respectively, $n$=5, and the number of patches used to train FCNNs was 5000*5*20.)

| Dataset | Methods | Dice | | | PPV | | | Sensitivity | | |
|---|---|---|---|---|---|---|---|---|---|---|
| | | complete | core | enhancing | complete | core | enhancing | complete | core | enhancing |
| Challenge | FCNN+CRF+post-process | 0.87 | 0.83 | 0.76 | 0.92 | 0.87 | 0.77 | 0.83 | 0.81 | 0.77 |
| | FCNN+CRF+post-1 | 0.87 | 0.83 | 0.76 | 0.92 | 0.87 | 0.77 | 0.83 | 0.81 | 0.77 |

|   | | | | | | | | | |
|---|---|---|---|---|---|---|---|---|---|
|   | FCNN+CRF+post-2 | 0.86 | 0.83 | 0.76 | 0.90 | 0.87 | 0.77 | 0.84 | 0.81 | 0.77 |
|   | FCNN+CRF+post-3 | 0.85 | 0.80 | 0.72 | 0.88 | 0.80 | 0.69 | 0.83 | 0.81 | 0.77 |
|   | FCNN+CRF+post-4 | 0.86 | 0.82 | 0.76 | 0.92 | 0.88 | 0.77 | 0.82 | 0.78 | 0.77 |
|   | FCNN+CRF+post-5 | 0.87 | 0.83 | 0.74 | 0.92 | 0.87 | 0.70 | 0.83 | 0.81 | 0.79 |
|   | FCNN+CRF+post-6 | 0.87 | 0.83 | 0.76 | 0.92 | 0.87 | 0.77 | 0.83 | 0.81 | 0.77 |
| Leaderboard | FCNN+CRF+post-process | 0.86 | 0.73 | 0.62 | 0.89 | 0.76 | 0.64 | 0.84 | 0.78 | 0.68 |
|   | FCNN+CRF+post-1 | 0.84 | 0.71 | 0.62 | 0.88 | 0.75 | 0.64 | 0.82 | 0.77 | 0.68 |
|   | FCNN+CRF+post-2 | 0.86 | 0.73 | 0.62 | 0.88 | 0.76 | 0.64 | 0.86 | 0.77 | 0.68 |
|   | FCNN+CRF+post-3 | 0.85 | 0.70 | 0.59 | 0.87 | 0.73 | 0.57 | 0.84 | 0.75 | 0.68 |
|   | FCNN+CRF+post-4 | 0.85 | 0.71 | 0.62 | 0.89 | 0.77 | 0.64 | 0.82 | 0.74 | 0.68 |
|   | FCNN+CRF+post-5 | 0.86 | 0.72 | 0.59 | 0.89 | 0.76 | 0.57 | 0.84 | 0.76 | 0.71 |
|   | FCNN+CRF+post-6 | 0.86 | 0.70 | 0.62 | 0.89 | 0.76 | 0.64 | 0.84 | 0.71 | 0.68 |

### *3.1.3. Evaluating the impact of image patch size*

We used different kernel sizes in all the pooling layers to train different FCNNs. The training image patch size changed with the kernel size, as shown in Fig. 4, while the number of parameters in FCNNs was unchanged. We evaluated the segmentation performance of our segmentation models with n=1,3,5, as summarized in Table 3. When n=1,3,5, the corresponding sizes of training patches are 21*21*3 (small input patch) and 41*41*3 (large input patch), 27*27*3 and 53*53*3, 33*33*3 and 65*65*3. Bar plots of the Dice of complete regions on the Challenge dataset with different training patch sizes are shown in Fig. 7. These segmentation results indicated that 1) a bigger patch provided more information and helped improve FCNNs' performance; 2) the CRF-RNN could reduce the performance differences caused by patch size as CRF-RNN could optimize the segmentation results according to the information in a whole image slice; and 3) the post-processing could further reduce the performance difference caused by patch size.

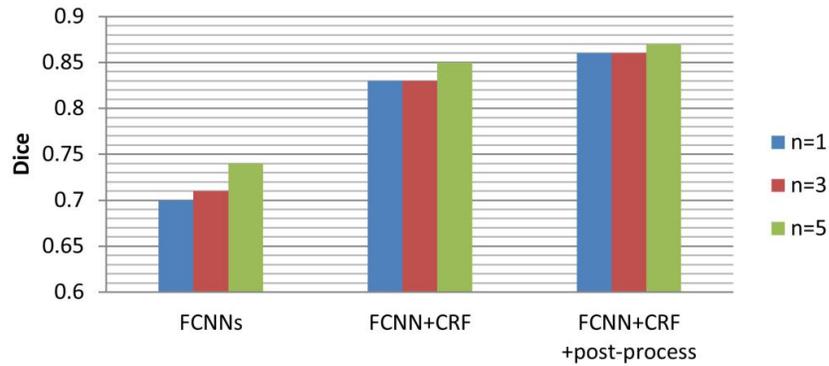

Fig.7. Bar plots for the Dice of complete regions on the Challenge dataset with different training patch sizes.

Table 3. Evaluation results of our segmentation model with n=1,2,3 (The number of patches used to train FCNNs was 5000*5*20.)

| Dataset | n | Methods | Dice | | | PPV | | | Sensitivity | | |
|---|---|---|---|---|---|---|---|---|---|---|---|
|   |   |   | complete | core | enhancing | complete | core | enhancing | complete | core | enhancing |
| Challenge | 1 | FCNNs | 0.70 | 0.62 | 0.64 | 0.57 | 0.50 | 0.55 | 0.94 | 0.87 | 0.79 |
|   |   | FCNN+CRF | 0.83 | 0.78 | 0.69 | 0.85 | 0.79 | 0.63 | 0.83 | 0.79 | 0.79 |
|   |   | FCNN+CRF+post-process | 0.86 | 0.83 | 0.76 | 0.92 | 0.88 | 0.77 | 0.82 | 0.80 | 0.77 |
|   | 3 | FCNNs | 0.71 | 0.67 | 0.65 | 0.58 | 0.56 | 0.57 | 0.95 | 0.86 | 0.78 |
|   |   | FCNN+CRF | 0.83 | 0.77 | 0.69 | 0.83 | 0.74 | 0.63 | 0.85 | 0.82 | 0.79 |
|   |   | FCNN+CRF+post-process | 0.86 | 0.82 | 0.76 | 0.91 | 0.84 | 0.78 | 0.84 | 0.82 | 0.76 |
|   | 5 | FCNNs | 0.74 | 0.72 | 0.67 | 0.62 | 0.63 | 0.60 | 0.94 | 0.86 | 0.77 |
|   |   | FCNN+CRF | 0.85 | 0.80 | 0.70 | 0.87 | 0.80 | 0.63 | 0.84 | 0.81 | 0.80 |
|   |   | FCNN+CRF+post-process | 0.87 | 0.83 | 0.76 | 0.92 | 0.87 | 0.77 | 0.83 | 0.81 | 0.77 |
| Leaderboard | 1 | FCNNs | 0.64 | 0.54 | 0.49 | 0.50 | 0.45 | 0.42 | 0.96 | 0.76 | 0.71 |

|   |   | FCNN+CRF | 0.82 | 0.65 | 0.57 | 0.83 | 0.70 | 0.51 | 0.85 | 0.69 | 0.72 |
|   |   | FCNN+CRF+post-process | 0.85 | 0.72 | 0.61 | 0.88 | 0.76 | 0.59 | 0.84 | 0.76 | 0.70 |
|   |   | FCNNs | 0.67 | 0.57 | 0.52 | 0.54 | 0.50 | 0.45 | 0.96 | 0.75 | 0.70 |
|   | 3 | FCNN+CRF | 0.83 | 0.66 | 0.57 | 0.82 | 0.67 | 0.51 | 0.87 | 0.72 | 0.71 |
|   |   | FCNN+CRF+post-process | 0.86 | 0.72 | 0.62 | 0.88 | 0.74 | 0.60 | 0.86 | 0.79 | 0.69 |
|   |   | FCNNs | 0.70 | 0.61 | 0.54 | 0.58 | 0.57 | 0.49 | 0.96 | 0.74 | 0.67 |
|   | 5 | FCNN+CRF | 0.83 | 0.66 | 0.57 | 0.85 | 0.71 | 0.50 | 0.85 | 0.69 | 0.71 |
|   |   | FCNN+CRF+post-process | 0.86 | 0.73 | 0.62 | 0.89 | 0.76 | 0.64 | 0.84 | 0.78 | 0.68 |

*3.1.4. Evaluating the impact of the number of image patches used to train FCNNs*

The BRATS 2013 training dataset contains 10 LGG and 20 HGG. In our experiments, we trained our segmentation model using the 20 HGG cases, and our model worked well for segmenting LGG cases. To investigate the impact of the number of training patches on the segmentation performance, we trained FCNNs with varied numbers of image patches. In particular, we sampled training imaging patches randomly from each subject and kept the number of training samples for different classes equal (5 classes in total, including normal tissue, necrosis, edema, non-enhancing core, and enhancing core). We generated 3 sets of image patches by sampling, consisting of 1000*5*20, 3000*5*20, and 5000*5*20 patches respectively, and used them to train different segmentation models. The evaluation results are summarized in in Table 4. Bar plots of the Dice of complete regions on the Challenge dataset with different numbers of training patches are shown in Fig. 8.

The results shown in Table 4 and Fig.8 indicated that the brain tumor segmentation accuracy of FCNNs increased with the increasing of the number of training patches. However, both CRFs and post-processing could reduce the performance difference.

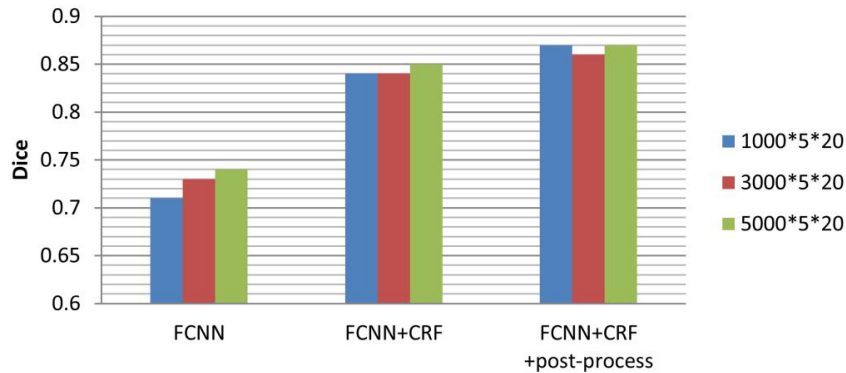

Fig.8. Bar plots for the Dice of complete regions on the Challenge dataset with different numbers of training image patches.

Table.4 Evaluation results of the segmentation models trained using different numbers of training image patches (the sizes of image patches used to train FCNNs were 33*33*3 and 65*65*3 respectively, $n$=5)

| Dataset | No. of patches | Methods | Dice | | | PPV | | | Sensitivity | | |
|---|---|---|---|---|---|---|---|---|---|---|---|
|   |   |   | complete | core | enhancing | complete | core | enhancing | complete | core | enhancing |
| Challenge | 1000*5*20 | FCNNs | 0.71 | 0.68 | 0.62 | 0.57 | 0.58 | 0.54 | 0.95 | 0.86 | 0.76 |
|   |   | FCNN+CRF | 0.84 | 0.78 | 0.68 | 0.84 | 0.78 | 0.61 | 0.85 | 0.81 | 0.79 |
|   |   | FCNN+CRF+post-process | 0.87 | 0.82 | 0.76 | 0.91 | 0.86 | 0.77 | 0.84 | 0.81 | 0.77 |
|   | 3000*5*20 | FCNNs | 0.73 | 0.70 | 0.66 | 0.60 | 0.61 | 0.59 | 0.94 | 0.86 | 0.76 |
|   |   | FCNN+CRF | 0.84 | 0.80 | 0.69 | 0.86 | 0.79 | 0.62 | 0.84 | 0.82 | 0.80 |
|   |   | FCNN+CRF+post-process | 0.86 | 0.83 | 0.77 | 0.92 | 0.86 | 0.77 | 0.83 | 0.82 | 0.77 |
|   | 5000*5*20 | FCNNs | 0.74 | 0.72 | 0.67 | 0.62 | 0.63 | 0.60 | 0.94 | 0.86 | 0.77 |
|   |   | FCNN+CRF | 0.85 | 0.80 | 0.70 | 0.87 | 0.80 | 0.63 | 0.84 | 0.81 | 0.80 |
|   |   | FCNN+CRF+post-process | 0.87 | 0.83 | 0.76 | 0.92 | 0.87 | 0.77 | 0.83 | 0.81 | 0.77 |

| Dataset | | Methods | | | | | | | | | |
|---|---|---|---|---|---|---|---|---|---|---|---|
| Learderboard | 1000*5*20 | FCNNs | 0.66 | 0.58 | 0.51 | 0.53 | 0.53 | 0.45 | 0.96 | 0.74 | 0.67 |
| | | FCNN+CRF | 0.83 | 0.65 | 0.56 | 0.82 | 0.71 | 0.51 | 0.86 | 0.69 | 0.70 |
| | | FCNN+CRF+post-process | 0.86 | 0.72 | 0.61 | 0.88 | 0.76 | 0.63 | 0.85 | 0.77 | 0.68 |
| | 3000*5*20 | FCNNs | 0.69 | 0.60 | 0.52 | 0.57 | 0.56 | 0.47 | 0.96 | 0.74 | 0.67 |
| | | FCNN+CRF | 0.83 | 0.65 | 0.56 | 0.84 | 0.70 | 0.49 | 0.85 | 0.69 | 0.71 |
| | | FCNN+CRF+post-process | 0.85 | 0.72 | 0.61 | 0.89 | 0.75 | 0.59 | 0.84 | 0.77 | 0.68 |
| | 5000*5*20 | FCNNs | 0.70 | 0.61 | 0.54 | 0.58 | 0.57 | 0.49 | 0.96 | 0.74 | 0.67 |
| | | FCNN+CRF | 0.83 | 0.66 | 0.57 | 0.85 | 0.71 | 0.50 | 0.85 | 0.69 | 0.71 |
| | | FCNN+CRF+post-process | 0.86 | 0.73 | 0.62 | 0.89 | 0.76 | 0.64 | 0.84 | 0.78 | 0.68 |

In summary, all the experimental results demonstrated that both CRFs and the post-processing method can narrow the performance difference caused by training patch sizes and training patch numbers, indicating that CRFs and the post-processing method might be able to narrow the performance difference caused by other training tricks. We will confirm this inference in our future work.

### 3.1.5. Performance Comparison between segmentation models built upon 4 and 3 imaging modalities

We also built a segmentation model using all available 4 imaging modalities, i.e., Flair, T1, T1c, and T2, and compared its segmentation performance with that of the segmentation model built upon 3 imaging modalities, i.e., Flair, T1c, and T2. The segmentation results of these segmentation models are summarized in Table 5. These results demonstrated that these two segmentation models achieved similar performance, indicating that a segmentation model built upon Flair, T1c, and T2 could achieve competitive performance as the model built upon 4 imaging modalities.

Table.5 Performance comparison of segmentation models built upon scans of 4 imaging modalities and 3 imaging modalities (the sizes of patches used to train FCNNs were 33*33*4 and 65*65*4, or 33*33*3 and 65*65*3 respectively, *n*=5, and the number of patches used to train FCNNs was 5000*5*20)

| Dataset | No. of modalities | Methods | Dice | | | PPV | | | Sensitivity | | |
|---|---|---|---|---|---|---|---|---|---|---|---|
| | | | complete | core | enhancing | complete | core | enhancing | complete | core | enhancing |
| Challenge | 4 | FCNNs | 0.74 | 0.72 | 0.67 | 0.61 | 0.64 | 0.60 | 0.95 | 0.86 | 0.77 |
| | | FCNN+CRF | 0.85 | 0.80 | 0.71 | 0.85 | 0.79 | 0.67 | 0.87 | 0.82 | 0.78 |
| | | FCNN+CRF+post-process | 0.87 | 0.83 | 0.76 | 0.91 | 0.85 | 0.78 | 0.84 | 0.82 | 0.76 |
| | 3 | FCNNs | 0.74 | 0.72 | 0.67 | 0.62 | 0.63 | 0.60 | 0.94 | 0.86 | 0.77 |
| | | FCNN+CRF | 0.85 | 0.80 | 0.70 | 0.87 | 0.80 | 0.63 | 0.84 | 0.81 | 0.80 |
| | | FCNN+CRF+post-process | 0.87 | 0.83 | 0.76 | 0.92 | 0.87 | 0.77 | 0.83 | 0.81 | 0.77 |
| Learderboard | 4 | FCNNs | 0.69 | 0.59 | 0.55 | 0.57 | 0.56 | 0.49 | 0.95 | 0.72 | 0.68 |
| | | FCNN+CRF | 0.82 | 0.64 | 0.58 | 0.82 | 0.70 | 0.55 | 0.85 | 0.67 | 0.68 |
| | | FCNN+CRF+post-process | 0.85 | 0.74 | 0.62 | 0.89 | 0.75 | 0.62 | 0.83 | 0.79 | 0.66 |
| | 3 | FCNNs | 0.70 | 0.61 | 0.54 | 0.58 | 0.57 | 0.49 | 0.96 | 0.74 | 0.67 |
| | | FCNN+CRF | 0.83 | 0.66 | 0.57 | 0.85 | 0.71 | 0.50 | 0.85 | 0.69 | 0.71 |
| | | FCNN+CRF+post-process | 0.86 | 0.73 | 0.62 | 0.89 | 0.76 | 0.64 | 0.84 | 0.78 | 0.68 |

### 3.1.6. Evaluation of different pre-processing strategies on the tumor segmentation

We preprocessed the imaging data using our robust deviation based intensity normalization and the standard deviation based intensity normalization [8], and then evaluated segmentation models built on them separately. As the results shown in Table 6 indicated, the robust deviation based intensity normalization could slightly improve the segmentation performance.

Table.6 Performance comparison of segmentation models built upon images normalized by the robust deviation and the standard deviation (the sizes of patches used to train FCNNs were 33*33*3 and 65*65*3 respectively, *n*=5, and the number of patches used to train FCNNs was 5000*5*20)

| Dataset | Deviation | Methods | Dice | | | PPV | | | Sensitivity | | |
|---|---|---|---|---|---|---|---|---|---|---|---|
| | | | complete | core | enhancing | complete | core | enhancing | complete | core | enhancing |
| | standard | FCNNs | 0.73 | 0.69 | 0.67 | 0.61 | 0.60 | 0.61 | 0.94 | 0.86 | 0.75 |
| | | FCNN+CRF | 0.84 | 0.80 | 0.71 | 0.87 | 0.79 | 0.68 | 0.82 | 0.82 | 0.75 |

| Challenge | | FCNN+CRF+post-process | 0.86 | 0.83 | 0.76 | 0.93 | 0.86 | 0.80 | 0.81 | 0.81 | 0.74 |
| | robust | FCNNs | 0.74 | 0.72 | 0.67 | 0.62 | 0.63 | 0.60 | 0.94 | 0.86 | 0.77 |
| | | FCNN+CRF | 0.85 | 0.80 | 0.70 | 0.87 | 0.80 | 0.63 | 0.84 | 0.81 | 0.80 |
| | | FCNN+CRF+post-process | 0.87 | 0.83 | 0.76 | 0.92 | 0.87 | 0.77 | 0.83 | 0.81 | 0.77 |
| Learderboard | standard | FCNNs | 0.69 | 0.60 | 0.54 | 0.57 | 0.55 | 0.50 | 0.97 | 0.75 | 0.67 |
| | | FCNN+CRF | 0.83 | 0.66 | 0.58 | 0.85 | 0.71 | 0.56 | 0.85 | 0.70 | 0.67 |
| | | FCNN+CRF+post-process | 0.86 | 0.73 | 0.61 | 0.89 | 0.75 | 0.66 | 0.84 | 0.78 | 0.66 |
| | robust | FCNNs | 0.70 | 0.61 | 0.54 | 0.58 | 0.57 | 0.49 | 0.96 | 0.74 | 0.67 |
| | | FCNN+CRF | 0.83 | 0.66 | 0.57 | 0.85 | 0.71 | 0.50 | 0.85 | 0.69 | 0.71 |
| | | FCNN+CRF+post-process | 0.86 | 0.73 | 0.62 | 0.89 | 0.76 | 0.64 | 0.84 | 0.78 | 0.68 |

*3.1.7. Evaluating the effectiveness of fusing the segmentation results gotten in three views*

We trained 3 segmentation models using patches and slices obtained in axial, coronal and sagittal views respectively. During testing, we used these three models to segment brain images from 3 views and got three segmentation results. The results of different views were fused and the evaluation results are shown in Table 7.

Evaluation results in Table 7 indicated that, for both Challenge and Leaderboard datasets, fusing the segmentation results typically led to better segmentation performance without the post-processing procedure. However, the improvement became insignificant after the post-processing procedure was applied to the segmentation results.

Table.7 Evaluations of segmentation results obtained in axial, coronal, sagittal views before and after post-processing, and evaluations of fusion results before and after post-processing (the sizes of patches used to train FCNNs were 33*33*3 and 65*65*3 respectively, *n*=5, and the number of patches used to train FCNNs was 5000*5*20)

| Dataset 2013 | Methods | | Dice | | | PPV | | | Sensitivity | | |
|---|---|---|---|---|---|---|---|---|---|---|---|
| | | | complete | core | enhancing | complete | core | enhancing | complete | core | enhancing |
| Challenge | FCNN+CRF | axial | **0.85** | 0.80 | 0.70 | **0.87** | 0.80 | 0.63 | 0.84 | 0.81 | 0.80 |
| | | coronal | 0.82 | 0.80 | 0.70 | 0.77 | 0.78 | 0.62 | **0.90** | **0.83** | **0.82** |
| | | sagittal | 0.83 | 0.80 | 0.71 | 0.80 | 0.79 | 0.64 | 0.87 | **0.83** | **0.82** |
| | fusing (FCNN+CRF) | | **0.85** | **0.83** | **0.74** | 0.85 | **0.84** | **0.68** | 0.88 | 0.82 | **0.82** |
| | FCNN+CRF+post-process | axial | 0.87 | 0.83 | 0.76 | **0.92** | **0.87** | **0.77** | 0.83 | 0.81 | 0.77 |
| | | coronal | 0.87 | 0.82 | 0.76 | 0.86 | 0.83 | 0.74 | **0.89** | **0.83** | 0.79 |
| | | sagittal | 0.87 | 0.82 | **0.77** | 0.88 | 0.83 | 0.75 | 0.86 | 0.82 | 0.79 |
| | fusing (FCNN+CRF)+post-process | | **0.88** | **0.84** | **0.77** | 0.90 | **0.87** | 0.76 | 0.86 | 0.82 | **0.80** |
| Learderboard | FCNN+CRF | axial | 0.83 | 0.66 | 0.57 | **0.85** | 0.71 | 0.50 | 0.85 | **0.69** | 0.71 |
| | | coronal | 0.80 | 0.65 | 0.56 | 0.75 | 0.69 | 0.50 | **0.90** | **0.69** | 0.70 |
| | | sagittal | 0.80 | 0.66 | 0.55 | 0.78 | 0.69 | 0.49 | 0.86 | **0.69** | 0.70 |
| | fusing (FCNN+CRF) | | **0.84** | **0.67** | **0.60** | 0.84 | **0.74** | **0.54** | 0.87 | 0.68 | **0.72** |
| | FCNN+CRF+post-process | axial | **0.86** | **0.73** | **0.62** | **0.89** | 0.76 | **0.64** | 0.84 | 0.78 | 0.68 |
| | | coronal | **0.86** | **0.73** | **0.62** | 0.86 | 0.73 | 0.60 | **0.89** | **0.79** | 0.67 |
| | | sagittal | 0.84 | 0.71 | 0.60 | 0.85 | 0.74 | 0.59 | 0.84 | 0.74 | 0.67 |
| | fusing (FCNN+CRF)+post-process | | **0.86** | **0.73** | **0.62** | **0.89** | **0.77** | 0.60 | 0.85 | 0.77 | **0.69** |

*3.1.8. Comparison with other methods*

Comparison results with other methods are summarized in Table 8. In particular, evaluation results of the top ranked methods participated in the BRATS 2013, shown on the BRATS 2013 website, are summarized in Table 8, along with the results of our method and other two state of art methods. Particularly, the method proposed by Sergio Pereira et al [31] ranked first on the Challenge dataset and second on the Leaderboard dataset right now, while our method ranked second on the Challenge dataset and first on the Leaderboard dataset right now. In general, it took 2-4 min for one of the three views of our method to segment one subject's imaging data. We do not have an accurate estimation of the training time for our segmentation models since we used a shared GPU server. On the shared GPU server, it took ~12 days to train our segmentation models.

Table. 8 Comparisons with other methods on BRATS 2013 dataset

| Dataset | Methods | Dice | | | Positive Predictive Value | | | Sensitivity | | |
|---|---|---|---|---|---|---|---|---|---|---|
| | | complete | core | enhancing | complete | core | enhancing | complete | core | enhancing |
| Challenge | Nick Tustison | 0.87 | 0.78 | 0.74 | 0.85 | 0.74 | 0.69 | 0.89 | **0.88** | **0.83** |
| | Raphael Meier | 0.82 | 0.73 | 0.69 | 0.76 | 0.78 | 0.71 | **0.92** | 0.72 | 0.73 |
| | Syed Reza | 0.83 | 0.72 | 0.72 | 0.82 | 0.81 | 0.70 | 0.86 | 0.69 | 0.76 |
| | Mohammad Havaei[30] | **0.88** | 0.79 | 0.73 | 0.89 | 0.79 | 0.68 | 0.87 | 0.79 | 0.80 |
| | Sergio Pereira[31] | **0.88** | 0.83 | **0.77** | 0.88 | **0.87** | 0.74 | 0.89 | 0.83 | 0.81 |
| | Our method(axial) | 0.87 | 0.83 | 0.76 | **0.92** | **0.87** | **0.77** | 0.83 | 0.81 | 0.77 |
| | Our method(coronal) | 0.87 | 0.82 | 0.76 | 0.86 | 0.83 | 0.74 | 0.89 | 0.83 | 0.79 |
| | Our method(sagittal) | 0.87 | 0.82 | **0.77** | 0.88 | 0.83 | 0.75 | 0.86 | 0.82 | 0.79 |
| | Our method(fusing) | **0.88** | **0.84** | **0.77** | 0.90 | **0.87** | 0.76 | 0.86 | 0.82 | 0.80 |
| Leaderboard | Nick Tustison | 0.79 | 0.65 | 0.53 | 0.83 | 0.70 | 0.51 | 0.81 | 0.73 | 0.66 |
| | Liang Zhao | 0.79 | 0.59 | 0.47 | 0.77 | 0.55 | 0.50 | 0.85 | 0.77 | 0.53 |
| | Raphael Meier | 0.72 | 0.60 | 0.53 | 0.65 | 0.62 | 0.48 | 0.88 | 0.69 | 0.64 |
| | Mohammad Havaei[30] | 0.84 | 0.71 | 0.57 | 0.88 | 0.79 | 0.54 | 0.84 | 0.72 | 0.68 |
| | Sergio Pereira[31] | 0.84 | 0.72 | **0.62** | 0.85 | **0.82** | 0.60 | 0.86 | 0.76 | 0.68 |
| | Our method(axial) | **0.86** | **0.73** | **0.62** | **0.89** | 0.76 | **0.64** | 0.84 | 0.78 | 0.68 |
| | Our method(coronal) | **0.86** | **0.73** | **0.62** | 0.86 | 0.73 | 0.60 | **0.89** | **0.79** | 0.67 |
| | Our method(sagittal) | 0.84 | 0.71 | 0.60 | 0.85 | 0.74 | 0.59 | 0.84 | 0.74 | 0.67 |
| | Our method(fusing) | **0.86** | **0.73** | **0.62** | **0.89** | 0.77 | 0.60 | 0.85 | 0.77 | **0.69** |

## 3.2. Segmentation performance on the BRATS 2015

The BRATS 2015 training dataset contains 54 LGG and 220 HGG, and its testing dataset contains 110 cases with unknown grades. All the training cases in BRATS 2013 are reused in BRATS 2015 as a subset of its training dataset. As what we have done in our experiments with BRATS 2013 dataset, we just used HGG training cases to train our segmentation models in this section. We extracted 1000*5 patches from each of 220 HGG to train FCNNs and initialed the CRF-RNN by the CRF-RNN trained by BRATS 2013 dataset. The whole network was fine-tuned using slices in BRATS 2013 training dataset, which is a subset of BRATS 2015 training dataset. The evaluation results of BRATS 2015 testing dataset are shown in Table 9, including evaluations of segmentation results that were segmented by the models trained by BRATS 2013 training dataset and evaluations of segmentation results that were segmented by the models trained in this section.

The results shown in Table. 9 indicated that fusing the segmentation results of multi-views could improve the segmentation accuracy. These results also indicated that a larger training dataset might improve the segmentation performance.

Table.9 Evaluation results of 110 testing cases in BRATS 2015 testing dataset (the sizes of image patches used to train FCNNs were 33*33*3 and 65*65*3 respectively, *n*=5)

| | Methods | | Dice | | | PPV | | | Sensitivity | | |
|---|---|---|---|---|---|---|---|---|---|---|---|
| | | | complete | core | enhancing | complete | core | enhancing | complete | core | enhancing |
| Models trained based on the BRATS 2013 training dataset | FCNN+CRF | axial | 0.77 | **0.56** | 0.52 | **0.85** | 0.71 | 0.46 | 0.74 | **0.53** | 0.67 |
| | | coronal | 0.74 | 0.53 | 0.48 | 0.70 | 0.69 | 0.44 | **0.84** | 0.50 | 0.65 |
| | | sagittal | 0.74 | 0.54 | 0.49 | 0.74 | 0.67 | 0.43 | 0.78 | 0.52 | 0.66 |
| | fusing (FCNN+CRF) | | **0.79** | **0.56** | **0.56** | 0.82 | **0.78** | **0.52** | 0.79 | 0.50 | **0.68** |
| | FCNN+CRF+post-process | axial | 0.78 | 0.64 | 0.58 | 0.87 | 0.75 | 0.57 | 0.73 | 0.61 | **0.65** |
| | | coronal | 0.78 | 0.61 | 0.57 | 0.77 | 0.68 | 0.58 | **0.84** | 0.63 | 0.63 |
| | | sagittal | 0.77 | 0.62 | 0.57 | 0.80 | 0.71 | 0.56 | 0.78 | 0.60 | 0.63 |
| | fusing (FCNN+CRF)+post-process | | **0.81** | **0.65** | **0.60** | 0.87 | **0.79** | **0.60** | 0.78 | 0.61 | **0.65** |

| | | | | | | | | | | |
|---|---|---|---|---|---|---|---|---|---|---|
| Models trained based on the BRATS 2015 training dataset | FCNN+CRF | axial | 0.78 | 0.64 | 0.54 | 0.78 | 0.76 | 0.48 | 0.81 | 0.62 | 0.71 |
| | | coronal | 0.77 | **0.66** | 0.56 | 0.73 | 0.73 | **0.52** | 0.86 | **0.67** | 0.67 |
| | | sagittal | 0.76 | 0.63 | 0.47 | 0.75 | 0.71 | 0.38 | 0.80 | 0.63 | **0.75** |
| | fusing (FCNN+CRF) | | **0.80** | **0.66** | **0.57** | **0.81** | **0.79** | 0.50 | 0.83 | 0.64 | 0.72 |
| | FCNN+CRF+post-process | axial | 0.80 | 0.68 | 0.61 | 0.82 | **0.75** | 0.59 | 0.81 | 0.71 | 0.68 |
| | | coronal | 0.80 | 0.69 | 0.61 | 0.77 | 0.70 | **0.61** | **0.87** | **0.76** | 0.65 |
| | | sagittal | 0.78 | 0.69 | 0.57 | 0.80 | 0.74 | 0.51 | 0.80 | 0.71 | **0.72** |
| | fusing (FCNN+CRF)+post-process | | **0.82** | **0.72** | **0.62** | **0.84** | **0.78** | 0.60 | 0.83 | 0.73 | 0.69 |

There were only 53 testing cases available during the BRATS 2015, but now there are 110 testing cases. Therefore, we are not able to directly compare our method with the methods that participated in the BRATS 2015. We are aware that K. Kamnitses et al [28] have published their evaluation results with 110 BRATS 2015 testing cases. The comparisons with K. Kamnitses et al's method are shown in Table 10.

Table. 10 Comparisons with other methods on BRATS 2015 dataset

| Dataset | Methods | Dice | | | Positive Predictive Value | | | Sensitivity | | |
|---|---|---|---|---|---|---|---|---|---|---|
| | | complete | core | enhancing | complete | core | enhancing | complete | core | enhancing |
| BRATS 2015 testing dataset | DeepMedic+CRF[28] | **0.847** | 0.67 | **0.629** | 0.85 | **0.848** | **0.634** | **0.876** | 0.607 | 0.662 |
| | Our method(axial) | 0.80 | 0.68 | 0.61 | 0.82 | 0.75 | 0.59 | 0.81 | 0.71 | 0.68 |
| | Our method(coronal) | 0.80 | 0.69 | 0.61 | 0.77 | 0.70 | 0.61 | 0.87 | **0.76** | 0.65 |
| | Our method(sagittal) | 0.78 | 0.69 | 0.57 | 0.80 | 0.74 | 0.51 | 0.80 | 0.71 | **0.72** |
| | Our method(fusing) | 0.82 | 0.72 | 0.62 | 0.84 | 0.78 | 0.60 | 0.83 | 0.73 | 0.69 |
| | Our FCNN(axial)+3D CRF | 0.84 | 0.72 | 0.62 | 0.88 | 0.75 | 0.62 | 0.82 | **0.76** | 0.67 |
| | Our FCNN(coronal)+3D CRF | 0.83 | 0.72 | 0.62 | 0.88 | 0.75 | 0.62 | 0.82 | 0.75 | 0.66 |
| | Our FCNN(sagittal)+3D CRF | 0.82 | 0.72 | 0.60 | 0.88 | 0.75 | 0.59 | 0.81 | **0.76** | 0.67 |
| | Our FCNN+3D CRF (fusing) | 0.84 | **0.73** | 0.62 | **0.89** | 0.76 | 0.63 | 0.82 | **0.76** | 0.67 |

### 3.3. Segmentation performance on the BRATS 2016

We also participated in the BRATS 2016. However, during the competition we just segmented brain images in axial view. Since the BRATS 2016 shares the same training dataset with BRATS 2015, we used the same segmentation models trained based on the BRATS 2015 training dataset. The BRATS 2015 training dataset has been pre-processed with rigid registration, bias field correction and skull stripping. However, the BRATS 2016 test dataset contains a number of unprocessed or partially pre-processed images [45], as shown in Fig. 9. To reduce false positives caused by incomplete pre-processing, apart from the post-processing steps described in Section 2.3.4, we manually placed rectangular bounding boxes around tumors in images and applied the post-processing step 2 to the segmentation results. Among 19 teams participated in the BRATS 2016, we ranked first on the multi-temporal evaluation. The ranking details of our method are shown in Table.11.

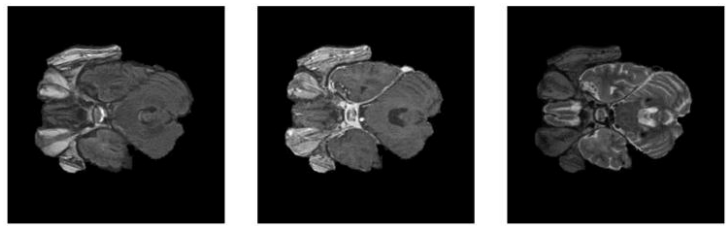

Fig.9 An example of partial skull stripping case in BRATS 2016 testing dataset. From left to right: Flair, T1c, T2

Table. 11 The ranking details of our method on different items on BRAST 2016 (including tie)

| Items | Tumor Segmentation | | | | | | Multi-temporal evaluation |
|---|---|---|---|---|---|---|---|
| | Dice | | | Hausdorff | | | |
| | complete | core | enhancing | complete | core | enhancing | |
| Ranking | 4 | 3 | 1 | 7 | 6 | 2 | 1 |

## 4. Discussions and Conclusion

In this study, we proposed a novel deep learning based brain tumor segmentation method by integrating Fully Convolutional Neural Networks (FCNNs) and Conditional Random Fields (CRFs) in a unified framework. This integrated model was designed to obtain tumor segmentation results with appearance and spatial consistency. In our method, we used CRF-RNN to implement CRFs [18], facilitating easy training of both FCNNs and CRFs as one deep network, rather than using CRFs as a post-processing step of FCNNs. Our integrated deep learning model was trained in 3 steps, using image patches and slices respectively. In the first step, image patches were used to train FCNNs. These image patches were randomly sampled from the training dataset and the same number of image patches for each class was used as training image patches, in order to avoid the data imbalance problem. In the second step, image slices were used to train the following CRF-RNN, with parameters of FCNNs fixed. In the third step, image slices were used to fine-tune the whole network. Particularly, we train 3 segmentation models using 2D image patches and slices obtained in axial, coronal and sagittal views respectively, and combine them to segment brain tumors using a voting based fusion strategy. Our experimental results also indicated that the integration of FCNNs and CRF-RNN could improve the segmentation robustness to parameters involved in the model training, such as image patch size and the number of training image patches. Our experimental results also demonstrated that a tumor segmentation model built upon Flair, T1c, and T2 scans achieved competitive performance as those built upon Flair, T1, T1c, and T2 scans.

We also proposed a simple pre-processing strategy and a simple post-processing strategy. We pre-processed each MR scan using N4ITK and intensity normalization, which normalized each MR image's intensity mainly by subtracting the gray-value of the highest frequency and dividing the robust deviation. The results shown in Fig. 1 and Fig. 2 demonstrated that the proposed intensity normalization method could make different MRI scans comparable, i.e., similar intensity values characterize similar brain tissues across scans. We post-processed the segmentation results by removing small 3D-connected regions and correcting false labels by simple thresholding method. Our experimental results have demonstrated that these strategies could improve the tumor segmentation performance.

Our method has achieved promising performance on the BRATS 2013 and BRATS 2015 testing dataset. Different from other top ranked methods, our method could achieve competitive performance with only 3 imaging modalities (Flair, T1c, and T2), rather than 4 (Flair, T1, T1c, and T2). We also participated in the BRATS 2016 and our method ranked first on its multi-temporal evaluation.

Our method is built upon 2D FCNNs and CRF-RNN to achieve computational efficiency. For training CRF-RNN and fine-tuning the integrated FCNNs and CRF-RNN, we use image slices as training data. However, in image slices, the numbers of pixels for different classes are different, which may worsen the segmentation performance of the trained network. To partially overcome the imbalanced training data problem, we trained CRF-RNN with the parameters of FCNNs were fixed so that the CRF-RNN are trained to optimize the appearance and spatial consistency of segmentation results. Such a strategy in conjunct with a fine-tuning of the whole network with a small learning rate improved the tumor segmentation performance. However, 2D CNNs are not equipped to take full advantage of 3D information of the MRI data [28, 29]. Our experimental results have demonstrated adopting 3D CRF as a post-processing step could improve the tumor segmentation performance. Our ongoing study is to build a fully 3D network to further improve the tumor segmentation performance.

## Acknowledgements

This work was supported in part by the National High Technology Research and Development Program of China (2015AA020504), the National Natural Science Foundation of China under Grant Nos. 61572499, 61421004, 61473296, and NIH grants EB022573, CA189523.